\documentclass[sigconf]{acmart}
\usepackage{amsmath,amsfonts}
\usepackage{textcomp}
\usepackage{url}
\usepackage{multirow}
\usepackage{mathtools}
\usepackage[linesnumbered,ruled,vlined]{algorithm2e}
\usepackage[noend]{algpseudocode}
\usepackage{soul}
\usepackage{subcaption}

\def\ceil#1{ \lceil #1 \rceil}
\AtBeginDocument{%
  \providecommand\BibTeX{{%
    \normalfont B\kern-0.5em{\scshape i\kern-0.25em b}\kern-0.8em\TeX}}}

\copyrightyear{2022}
\acmYear{2022}
\setcopyright{rightsretained}
\acmConference[ISLPED '22]{ACM/IEEE International Symposium on Low Power Electronics and Design}{August 1--3, 2022}{Boston, MA, USA}
\acmBooktitle{ACM/IEEE International Symposium on Low Power Electronics and Design (ISLPED '22), August 1--3, 2022, Boston, MA, USA}
\acmDOI{10.1145/3531437.3539715}
\acmISBN{978-1-4503-9354-6/22/08}

\begin{document}
%\pagestyle{plain}

\title{Sparse Periodic Systolic Dataflow for Lowering Latency and Power Dissipation of Convolutional Neural Network Accelerators}


\author{Jung Hwan Heo}
\authornote{Both authors contributed equally to this research.}
\affiliation{
  \institution{University of Southern California}
  \city{Los Angeles}
  \state{California}
  \country{USA}
}
\email{johnheo@usc.edu}

\author{Arash Fayyazi}
\authornotemark[1]
%\author{Amirhossein Esmaili}
%\email{webmaster@marysville-ohio.com}
\affiliation{%
  \institution{University of Southern California}
  \city{Los Angeles}
  \state{California}
  \country{USA}
}
\email{fayyazi@usc.edu}

\author{Amirhossein Esmaili}
\affiliation{%
  \institution{University of Southern California}
  \city{Los Angeles}
  \state{California}
  \country{USA}
}
\email{esmailid@usc.edu}

\author{Massoud Pedram}
\affiliation{%
  \institution{University of Southern California}
  \city{Los Angeles}
  \state{California}
  \country{USA}
}
\email{pedram@usc.edu}


\begin{abstract}
{This paper introduces the sparse periodic systolic (SPS) dataflow, which advances the state-of-the-art hardware accelerator for supporting lightweight neural networks. Specifically, the SPS dataflow enables a novel hardware design approach unlocked by an emergent pruning scheme, periodic pattern-based sparsity (PPS). By exploiting the regularity of PPS, our sparsity-aware compiler optimally reorders the weights and uses a simple indexing unit in hardware to create matches between the weights and activations. Through the compiler-hardware codesign, SPS dataflow enjoys higher degrees of parallelism while being free of the high indexing overhead and without model accuracy loss. Evaluated on popular benchmarks such as VGG and ResNet, the SPS dataflow and accompanying neural network compiler outperform prior work in convolutional neural network (CNN) accelerator designs targeting FPGA devices. Against other sparsity-supporting weight storage formats, SPS results in 4.49$\times$ energy efficiency gain while lowering storage requirements by 3.67$\times$ for total weight storage (non-pruned weights plus indexing) and 22,044$\times$ for indexing memory.}
\end{abstract}

%   <concept>
%       <concept_id>10010583.10010786</concept_id>
%       <concept_desc>Hardware~Emerging technologies</concept_desc>
%       <concept_significance>500</concept_significance>
%       </concept>
%     <concept>
%       <concept_id>10010147.10010257</concept_id>
%       <concept_desc>Computing methodologies~Machine learning</concept_desc>
%       <concept_significance>500</concept_significance>
%       </concept>
%  </ccs2012>
% \ccsdesc[500]{Hardware~Emerging technologies}
% \ccsdesc[500]{Computing methodologies~Machine learning}


% \ccsdesc[500]{Hardware~Power and energy}


\maketitle

\vspace*{-6pt}

\section{Introduction}
Convolutional Neural Networks (CNNs) exhibit great performance in many computer vision applications such as image classification, object recognition, and scene labeling \cite{DBLP:conf/nips/KrizhevskySH12}. However, the high performance of deep CNNs is achieved at the cost of high computation. This makes it challenging for networks to be deployed to resource-constrained edge devices with strict storage and energy limits. Therefore, developing CNN architectures with reduced computation and storage costs is of great importance. At the algorithmic level, methods such as weight quantization \cite{NazemiFEKSP21}, weight pruning \cite{DBLP:journals/corr/HanPTD15, DBLP:conf/nips/DingDZGHL19}, and knowledge distillation \cite{DBLP:journals/corr/HintonVD15} have gained recent popularity.   

In particular, weight pruning is a widely practiced approach for reducing the memory footprint and computational cost of neural networks. By removing redundant weights of a network that does not harm the model accuracy, the model is compressed from a dense to a sparse computational graph. With the progress in weight pruning methods, pattern-based pruning \cite{patdnn,DBLP:journals/tc/KunduNPCB20} has emerged as a promising avenue that seeks to find a sweet spot between the two conventional pruning schemes: 1) structured pruning \cite{DBLP:conf/nips/WenWWCL16} which has high regularity and is hardware-friendly, but susceptible to accuracy degradation; 2) unstructured pruning which retains high accuracy, but suffers from large hardware overhead to manage irregular weight indices. Pattern-based pruning method compromises between these two pruning schemes by enforcing a semi-structured level of regularity through pre-defined patterns. This ameliorates the hardware overhead compared to unstructured pruning, but it still necessitates a series of auxiliary buffers to manage a unique set of indexing scenarios with the pattern-based approach. At its core, hardware overhead caused by indexing sparse weights manifests a fundamental design limitation for the accelerator to further optimize latency, power, and memory requirements. 

In this paper, we advance the state-of-the-art in sparse neural network accelerator design by exploiting the concept of periodicity in pattern-based pruning for the first time in hardware. Prior art \cite{DBLP:journals/tc/KunduNPCB20} mainly explores the software stack of the periodic pattern-based pruning approach and demonstrate that added periodicity has negligible accuracy loss. Here, we observe periodicity as an opportunity in hardware to avoid indexing overhead with its added regularity. We first present our compiler that reorders the weights according to the periodicity, optimizing for maximum parallelism. Then we present sparse periodic systolic (SPS) dataflow that computes convolutions in a systolic array of processing elements, commonly seen in Field Programmable Gate Arrays (FPGAs). Then, a dedicated indexing method is introduced in hardware to fetch the pre-defined locations of nonzero weight indices using significantly smaller memory requirements.
The main contributions of this paper are summarized as follows:
\vspace*{-6pt}
\begin{itemize}
    \item We present a novel  SPS dataflow that exploits the \emph{periodic pattern-based sparsity} in neural networks to achieve an FPGA-friendly architecture.
    \item Using a compiler tailored to the SPS dataflow, we effectively solve the long-standing indexing overhead problem for unstructured pruning. We co-design the period-pattern-weight (PPW) compact storage format and the corresponding architecture to efficiently fetch weights and activations.
    \item We perform the next layer reordering (NLR) optimization method enabled by the periodic pattern-based design to further reduce data movement cost in between layers.
\end{itemize}

\section{Preliminaries and Background}
This section includes background on deep neural network (DNN) processing and details the \emph{periodic pattern-based pruning} method, which is the weight reduction technique used for DNN compression in this paper.

\subsection{DNN processing and Compilers}
A convolutional layer receives input feature feature maps (IFMs) of size $w_\mathrm{in} \times h_\mathrm{in} \times c_\mathrm{in}$ and convolves them with $c_\mathrm{out}$ different filters, each filter of size $w_\mathrm{k} \times h_\mathrm{k} \times c_\mathrm{in}$ to generate output feature maps (OFMs) of size $w_\mathrm{out} \times h_\mathrm{out} \times c_\mathrm{out}$. Here, $w_\mathrm{x}$, $h_\mathrm{x}$, $c_\mathrm{x}$ represent width, height, and depth of tensor $x$, which can represent the 3D input/output feature map. The IFMs for the next convolutional layer are equivalent of the current convolutional layer's OFMs. Such computations can be represented by a six-level nested loop (seven-level nested loops when considering iteration over images in a mini-batch), i.e., loops over $w_\mathrm{out}, h_\mathrm{out}, c_\mathrm{out}, w_\mathrm{k}, h_\mathrm{k},$ and $c_\mathrm{in}$. Also known as a computational block, these nested loops characterize the computational flow for a convolutional layer in CNNs.

\subsection{Periodic Pattern-based Sparsity (PPS)}
Pattern-based sparsity will first be explained before the introduction of periodicity. In pattern-based sparsity, a pattern is defined as a pre-defined 2D kernel that constrains the locations of nonzero entries, also referenced as a kernel variant (KV). Thus, any given kernel of a 3D filter can be classified as one of the KVs, since the locations of the weights are strictly assigned to form a pattern while pruning. The number of nonzero entries (the kernel support) in a $w_\mathrm{k} \times h_\mathrm{k}$ kernel is also referred to kernel support size (KSS). KSS is fixed for all patterns to support high regularity, which reduces workload imbalance between processing elements (PEs) in the systolic array. Unlike unstructured pruning that prunes at the granularity of individual weights, pattern-based pruning prunes at the granularity of patterns, which adds regularity yet less flexibility. The regularity helps with achieving higher hardware performance, but less flexibility poses a relative challenge in retaining accuracy.
\begin{figure}[t]
  \centering
  \includegraphics[width=0.4\textwidth]{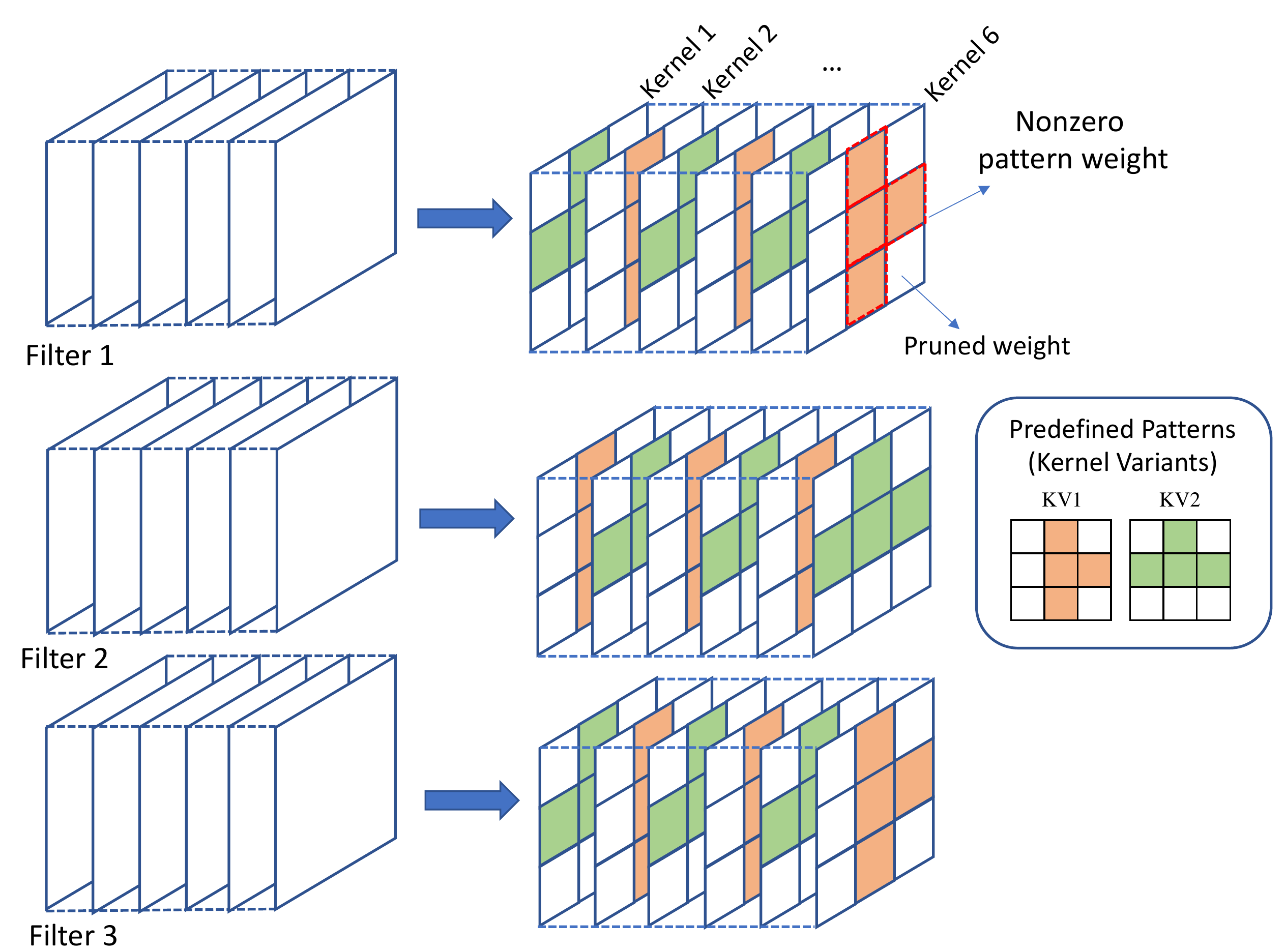}
  \caption{Illustration of PPS with params KSS=4 and P=2.}
  \label{fig:ppw-example}
\end{figure}

Periodic pattern-based sparsity is an extension of the pattern-based sparsity, where the concept of periodicity constrains the sequence in which the patterns occur in a given filter. Fig. \ref{fig:ppw-example} illustrates an example of PPS. Rotation of patterns (or KVs) due to periodicity occurs in two directions, across the kernels and across the filters. Each KV appears in a repeating sequence of [KV1, KV2, KV1, ...] for the first filter. Such sequence of filters with a unique initial KV is denoted as a filter variant (FV). For the second FV, it will begin with KV2 in a rotating sequence of [KV2, KV1, KV2, ...]. Having rotations across both channel and filter dimension adds flexibility to improve network accuracy.

A key insight here is the simplicity at which the KVs can be indexed in an any arbitrary filter. Thanks to the modulo rotation that occurs with an interval of periodicity (P), the burden of storing the location of each KV (or pattern) can be reduced to a single scalar value P, which is also the number of KVs. This means the weights associated with each KV can be accessed by P with minimal overhead. Thus, every KV of same type can be indexed by iterating across the filter with offset P, which is much simpler than iterating over the indices of each KV type and its respective location that irregularly occurs across the filter.

\begin{figure*}[t]
  \centering
  \includegraphics[width=0.75\textwidth]{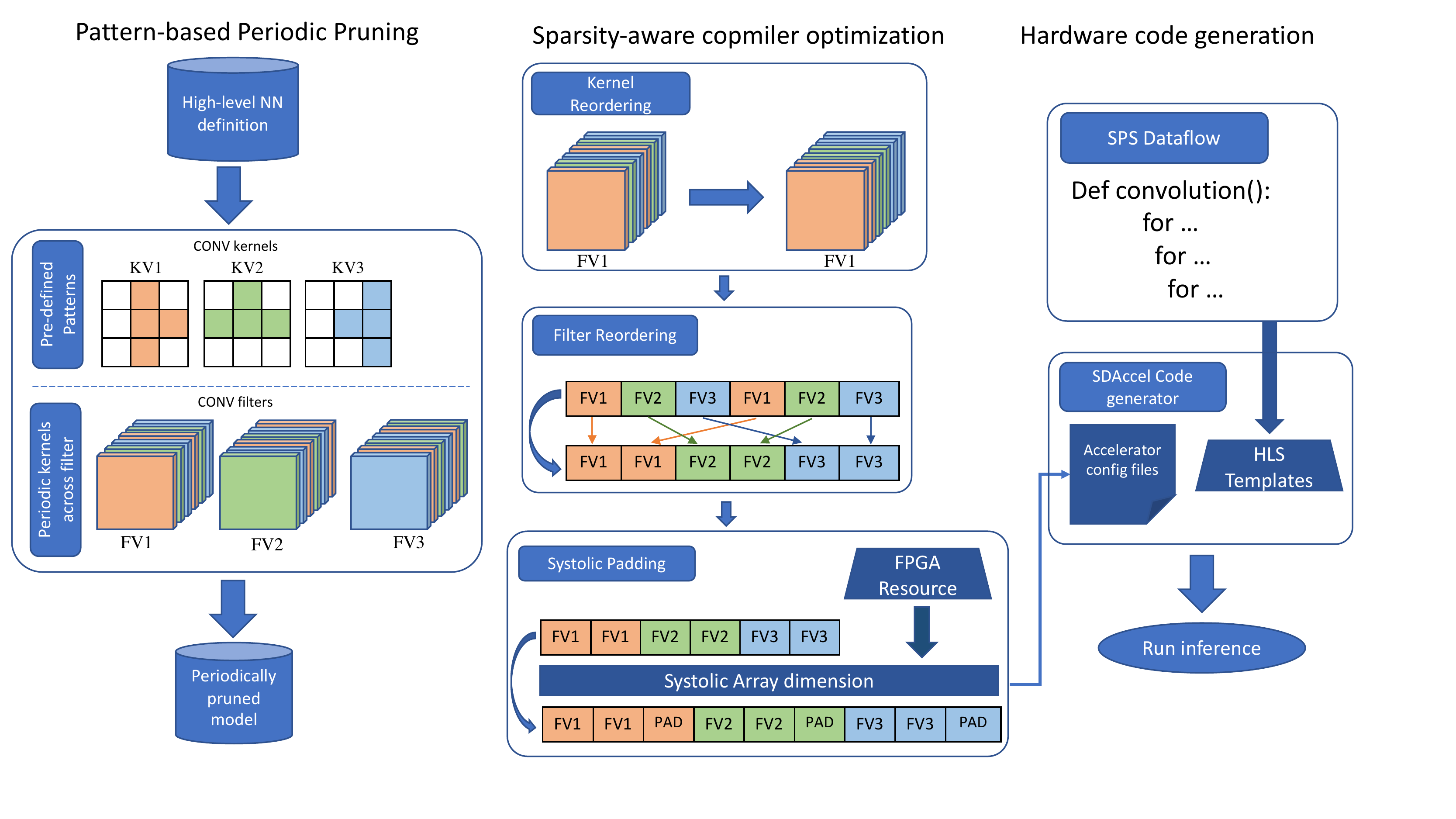}
  \caption{Overall flow of the SPS Acceleration Framework.}
  \label{fig:overall-flow}
  %\vspace{-10pt}
\end{figure*}

\section{Overall Flow}
Given the golden opportunity to design a hardware that does not suffer from indexing overhead while preserving the network accuracy, we propose a novel end-to-end FPGA-friendly DNN acceleration framework that can fully exploit the new \emph{periodic pattern-based} dimension in its dataflow.
%The solution to the overhead problem is to emulate the dense dataflow, where weights are efficiently fetched without extra indexing systems. 
%Through an architecture-first approach, we develop a highly tailored dataflow with high parallelism, essentially imitating the indexing-free dense dataflow. 
Fig. \ref{fig:overall-flow} shows our acceleration framework that consists of three stages. First is the model pruning stage, where we employ pattern-based periodic pruning method developed by \cite{DBLP:journals/tc/KunduNPCB20}. Second is the sparsity-aware compiler optimization (cf. Section \ref{sec:compiler}) that performs a series of periodicity-driven weight reshaping operations. The compiler provides a maximum degree of flexibility for model compression parameters, such as pattern shapes, P, and KSS. Systolic padding is also applied to maintain the parallelism that occurs across the two dimensions of the systolic array. Last is the sparsity-aware architecture (cf. Section \ref{sec:arch}), where the input matching Unit (IMU) is designed to facilitate the SPS dataflow in an FPGA-friendly hardware architecture.

\section{Compiler Tailored to PPS}
\label{sec:compiler}
Once the model is trained using pattern-based periodic pruning, sparse weights must be stored in an efficient format. Otherwise, the indexing overhead will cause the expected benefits of pruning to be lost. As such, this section introduces a compiler that shapes the initial weight tensor to a compact format that we call period-pattern-weight (PPW) format. From the initial 4D weight tensor that is periodically pattern-pruned with periodicity P and KSS number of nonzero weights per KV, we apply a series of transformations to arrive at the sparsity parameters described below in Table \ref{tab:notations}.

\subsection{Sparsity-aware Optimizations}
Key challenges of hardware acceleration for unstructured pruning can be reduced to heavy control-flow instructions, as well as thread divergence and load imbalance \cite{DBLP:hwaccel-survey}. These are largely solved by promoting parallelism during the dataflow. Grouping filters with similar kernel sequences achieves better \emph{inter-thread} parallelization, while grouping same patterns within a filter improves \emph{intra-thread} parallelization \cite{patdnn}. Taken together, these provide a key insight as to how the patterns should be rearranged by the compiler to maximize parallelism.

\begin{figure}[b]
  \centering
  \includegraphics[width=.85\columnwidth]{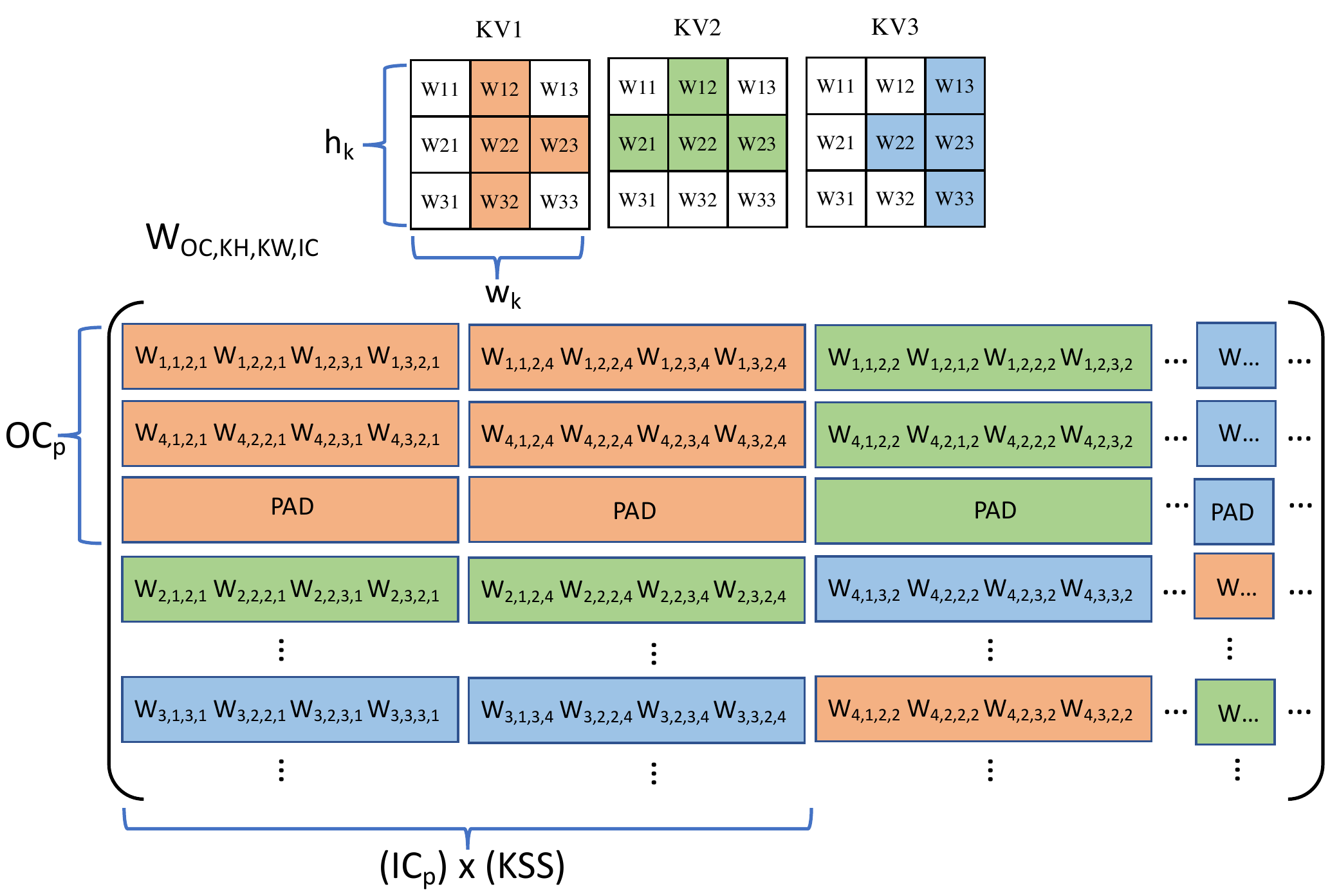}
  \caption{Overview of the PPW storage format.} 
  \label{fig:compiler-flow}
  \vspace{-2pt}
\end{figure}

\textbf{Kernel and Filter Reordering: } Kernel and filter reorderings both group the KVs according to P. A running example is illustrated in Fig. \ref{fig:overall-flow} where it shows a grouping of three distinct KVs together. Periodicity of three results in a first group KV1 that increments in multiples of three with the offset of one, which is the KV number. Similarly, KV2 increments in P with offset of 2 and so on. This results in 3 arrays of input channels (ICs) in the form of [1+P*0, 1+P*1, 1+P*2, ...], [2+P*0, 2+P*1, ...], and [3+P*0, 3+P*1, ...] assuming 1-based indexing. Each array has a size of $\ceil{c_{in}/P}$ denoted by $IC_p$.

Once the same KVs are grouped together, the same \emph{sequences of kernels} are grouped together as well. Each unique sequence produces a filter variant (FV), where in our example equates to three groups of output channels (OC) being produced with each group having a size of $\ceil{c_{out}/P}$, denoted by $OC_p$ (see Fig. \ref{fig:compiler-flow}). 

\begin{table}[b]
\centering
\caption{\small Parameters used in the proposed compiler.} 
\label{tab:notations}
\resizebox{\columnwidth}{!}{\begin{tabular}{c|c|l}
\textbf{Symbol}  & \textbf{Calculation} & \textbf{Description} \\
\hline\hline
%$P$ & Given & \# of patterns in the current CONV layer\\ 
%$KSS$ & Given & \# of nonzero weights per pattern\\
$W_{NUM}$ & $P \times KSS$ & total \# of predefined nonzero weights\\
$IC_p$ & $\ceil{c_\mathrm{in}/P}$ & \# of input channels per group\\
$INC_p$ & $\ceil{c_\mathrm{cin}/P/sys_w}$ & \# of input channels tiled iterations per group\\
$OC_p$ & $\ceil{c_\mathrm{out}/P}$ & \# of output channels per group\\
$ONC_p$ & $\ceil{c_\mathrm{out}/P/sys_h}$ & \# of output channels tiled iterations per group\\
\end{tabular}}
\end{table}

\textbf{Systolic Padding: }
To support the SPS dataflow (Section \ref{sec:sps-dataflow}) that concurrently processes groups of input channels $IC_p$ and output channels $OC_p$ together, padding is applied when either of these groups doesn't fully populate the systolic array resources.
It is worth noting that padding is usually interpreted as a wasteful operation and should be avoided whenever possible, for example by pipelining. %More advanced optimizations may be applied by computing two different patterns if there is excessive padding. 
However, in our experience, the number of patterns required to achieve high performance is 6-8. This means that if the dimension of the weight matrix ($IC_p$, $OC_p$) is divisible by ($sys_w$, $sys_h$), no padding will be necessary. Fortunately, most well-known networks such as VGG and ResNets have weight dimensions in multiples of 8 with popular ones being 128, 256, and 512. Since the quotient values can be used as reconfigurable systolic array dimensions, the cost incurred from padding is minimal.

\let\oldnl\nl% Store \nl in \oldnl
\newcommand{\nonl}{\renewcommand{\nl}{\let\nl\oldnl}}% Remove line number for one line
\newlength{\textfloatsepsave} \setlength{\textfloatsepsave}{\textfloatsep}
\setlength{\textfloatsep}{0pt}
\newcommand{\plusplus}{\raisebox{.4\height}{\scalebox{.9}{++}}}
\begin{algorithm}[t]
\caption{The Sparse Periodic Systolic Dataflow} \label{alg.mapping}
\small
\SetAlgoLined
\KwIn{$W^{oc,ic}_{kh,kw}$: Nonzero weights;
        \hspace{2pt}$A^{ic}_{kh,kw}$: Input activations;\\
        \hspace{25pt}$w_\mathrm{k} \times h_\mathrm{k}$: Kernel Size; PS[][]: Partial Sum register in PE \\
        \hspace{25pt}$h_\mathrm{out} \times w_\mathrm{out} \times c_\mathrm{out}$: Output Feature Map (OFM) Size;
}
\KwOut{Result stored in OFM}
\For{$oh = 0; oh<h_{out}; oh\plusplus$}{
    \For{$ow = 0; ow<w_{out}; ow\plusplus$}{
    \nonl // begin convolution \\
        \For{$g = 0; g<P; g\plusplus$}{
            \For{$cc = 0; cc<ONC_{p}; cc\plusplus$}{
                \For{$kv = 0; kv<P; kv\plusplus$}{
                    \For{$w = 0; w<KSS; w\plusplus$}{
                    \nonl // read Weight Index Buffer (kh, kw) \\
                        \For{$rr = 0; rr<INC_{p}; rr\plusplus$}{
                            \For{$i = 0; i<w_{sys}; i\plusplus$}{
                            \#pragma unroll(i) \\
                                \For{$j = 0; j<h_{sys}; j\plusplus$}{
                                \#pragma unroll(j) \\
                                PS[j][i]+=$W^{j,i}_{kh,kw}*A^{i}_{kh,kw}$
                                }
                            }
                        }
                    }
                }
                \nonl // Tree Adder to add Partial Sums across $w_{sys}$ \\
                \nonl // Accumulate partial OFM in Output Buffer
            }
            \nonl // Store resulting OFM from the buffer to DRAM
        }
    }
}
\end{algorithm}

\section{The Sparse Periodic Systolic Dataflow}
\label{sec:sps-dataflow}
The SPS Dataflow guides compact weights to be run in the FPGA-friendly systolic architecture. It has two major functionalities: 1) matching the PPW weight tensor with the corresponding activations and 2) tiling across the $INC_p$ and $ONC_p$ by concurrently executing all MAC operations in the systolic array. 

First, the \emph{temporal} component of the SPS Dataflow may be understood by looking at a single PE unit in the 2D systolic array of the hardware accelerator. As Section \ref{sec:arch} describes, weights are stored in the BRAMs in each PE and the output of the associated MAC unit is stored in a partial sum register inside the PE. Thus, the dataflow is a combination of weight stationary and output stationary dataflow. This order maximizes the reuse of weights as well as outputs, while paying some cost to stream the activations to the computation units. 

The \emph{spatial} component of the SPS Dataflow is mapped to match the dimensions of the systolic array. The compiler has already grouped the input and output channels according to weight patterns, and two additional inner loop nests (i and j iterators in Algorithm 1) further blocks out the subgroup within $IC_p$ by $sys_w$ and $OC_p$ by $sys_h$, resulting in $INC_p$ and $ONC_p$, respectively.

The crux of the SPS dataflow lies in the simplicity of decoding the compressed PPW format to fetch the corresponding input activations. Many state-of-the-art sparsity-supported accelerators use storage formats such as the coordinate list (COO), compressed sparse row (CSR), and compressed sparse column (CSC) \cite{cheng2014professional}, where the storage requirement for nonzero indexing polynomially increases with the network size. However, the storage for PPW is network architecture-agnostic, meaning it can stand on a constant storage requirement only dependent on pruning parameters P and KSS, regardless of how deep or wide the network is. 
% A small figure would be nice here

The method to facilitate the MAC operation indexing between activation and the weight is as follows. We create two indexing buffers of size $W_{NUM}$, each responsible for storing the two spatial dimensions of a kernel, $h_\mathrm{k}$ and $w_\mathrm{k}$. Similar to the COO format, each weight can be indexed by a single iterator that fetches the height and width of the weight inside the kernel. Thanks to the highly regular occurrence of the patterns, each weight in a given group, kernel number, and nonzero weight number can be calculated by $((g+kv)*KSS +w)\%W_{NUM}$. The modulus operation wraps the buffer so it continues the periodically occurring patterns.

%\textbf{Next Layer Reordering (NLR):}
\subsection{Next Layer Reordering }
After the convolution operation is completed, the results of the MAC operations are stored. Due to compiler's reordering, the unnatural ordering of output channels produces results in increments of P. A naive solution is to simply sort it to a natural order (increasing from channel 0 to $c_\mathrm{cout}-1$. Yet, this causes a nontrivial amount of data movement from scanning and reordering the entire output channel for \emph{every layer transition} and also accessing the buffer that stores the the sorting indices. 

Here, we observe that SPS dataflow produces the OFM with an increment of P. Therein, the compiler can expect the channels to be grouped in certain strides of P and \emph{proactively} reorder the channels for the next layer so that it matches the channel ordering of incoming activations. As such, NLR can save the total execution time and energy efficiency from reordering after each convolutional layer, effectively imitating the dense dataflow where such channel indexing problems do not occur.

\begin{figure}[b]
  \centering
  \includegraphics[width=0.8\columnwidth]{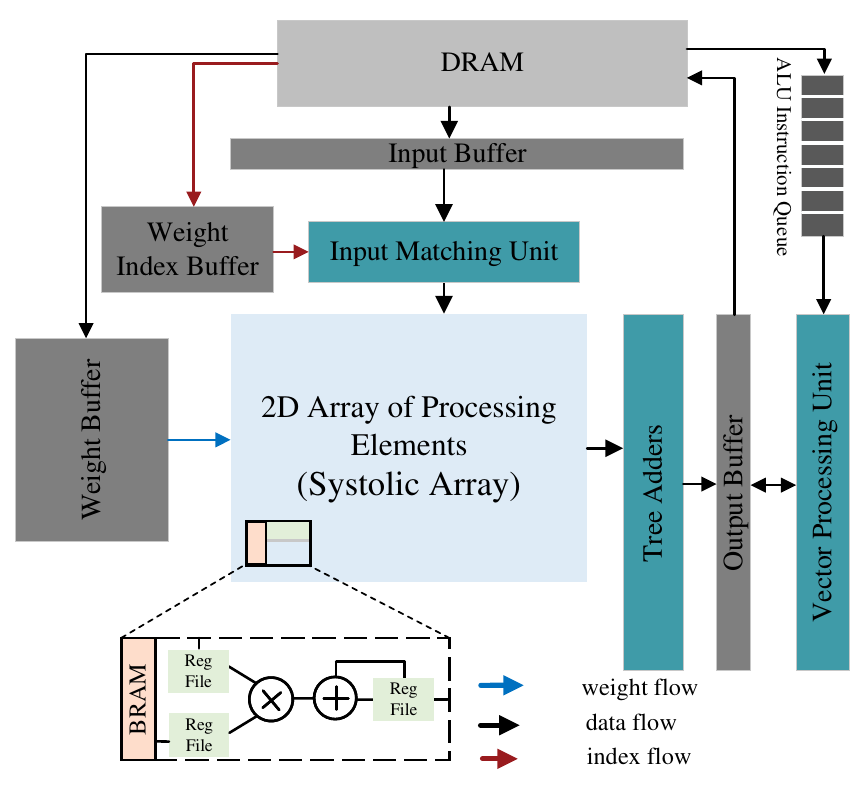}
  \caption{Overview of systolic array accelerator design.}
  \label{fig:arch}
\end{figure}

\section{Proposed Periodic Sparsity Architecture}
\label{sec:arch}
In this section, we describe our FPGA-tailored architecture customized for the proposed periodic pattern-based sparsity.

The accelerator contains (i) a 2D array of PEs (systolic array) which is responsible for executing the MAC operations associated with the convolution operation, (ii) a memory hierarchy feeding data to the said array, which consists of register files, on-chip memory (Block RAMs on FPGA devices), and external off-chip memory (DRAM), and (iii) an input matching unit (IMU) that reads the nonzero weight indices from the Weight Index Buffer and matches with the input feature maps. The systolic array is followed by a vector processing unit, which includes multiple ALUs that conduct neural network operations such as nonlinear activation functions and maximum pooling, as illustrated in Fig. \ref{fig:arch}. 

The available hardware resources in an FPGA device, such as digital signal processing units (DSPs), Configurable Logic Blocks (CLBs) that contain several look-up tables (LUTs), and Block RAMs (BRAMs) are placed as resource groups in a column-wise manner. Consequently, the all resources are uniformly distributed on the FPGA chip, and one should place data that is used by a DSP in a BRAM that is physically close to the DSP. Hence, A PE in our design comprises one DSP and its adjacent BRAM. We also use some of the CLBs as distributed memories to store indices of non-zero weights in KVs. Note that these indices are low precision (e.g., 4 bits for a 2D kernel with size of $3 \times 3$ which is common in well-known computer vision models).

The IFMs are initially cached in an input buffer, then passed through the IMU to skip pruned weights, and sequentially transmitted onto the first row of PEs in the systolic array. In addition, input data is simply shifted into the PE array and between nearby PEs on the same row of the systolic array. This technique does away with the need for global interconnections between the input buffers and all PEs and the costly multiplexers. We also bring the indices associated with KVs in parallel with weight fetching. This is feasible since input data, weights, and indices are stored in separate off-chip memory banks in the target FPGA board and are thus simultaneously accessible. Finally, the registered partial sum results that reside in the PEs of one row are passed to the adder tree to conduct the required summation and generate the final OFM value when all computations for one OFM are completed.

\section{Experimental Results}
\label{sec:exp}
In this section, we assess the storage required by the proposed PPW format compared to popular sparsity-supporting formats and present the hardware utilization of our accelerator as well as its energy consumption comparison to state-of-the-art accelerators.
\subsection{Experimental Configuration}
% reference ESCA and F2N2 describing the board the evaluation environment. 
For the storage format experiments, 8 bit unsigned integers are used to calculate the weight storage format. For a fair comparison, the connectivity pruning that allows higher weight compression for FKW format is recognized during calculation. Sparsity constants of KSS=2 and P=8 is used for all PPW calculations, as we validate the model accuracy (91.2\%) \cite{DBLP:journals/tc/KunduNPCB20} that has less than 1\% accuracy degradation compared to the non-pruned version. 
For evaluating hardware performance, we targeted a Xilinx VU9P FPGA using the AWS EC2 F1 instance. We implemented it on Xilinx Virtex UltraScale+ FPGA board %(xcvu9p-fsgd2104-2-i or otherwise VU9P) 
using Vivado HLS design suite 2019.1. We evaluate our SPS Dataflow on a VGG16 architecture on the CIFAR-10 dataset.

\subsection{Storage Comparison}
\label{sub:storage}

\begin{figure}[t]
    \centering
    \begin{minipage}{0.48\columnwidth}
        \centering
        \includegraphics[width=\textwidth]{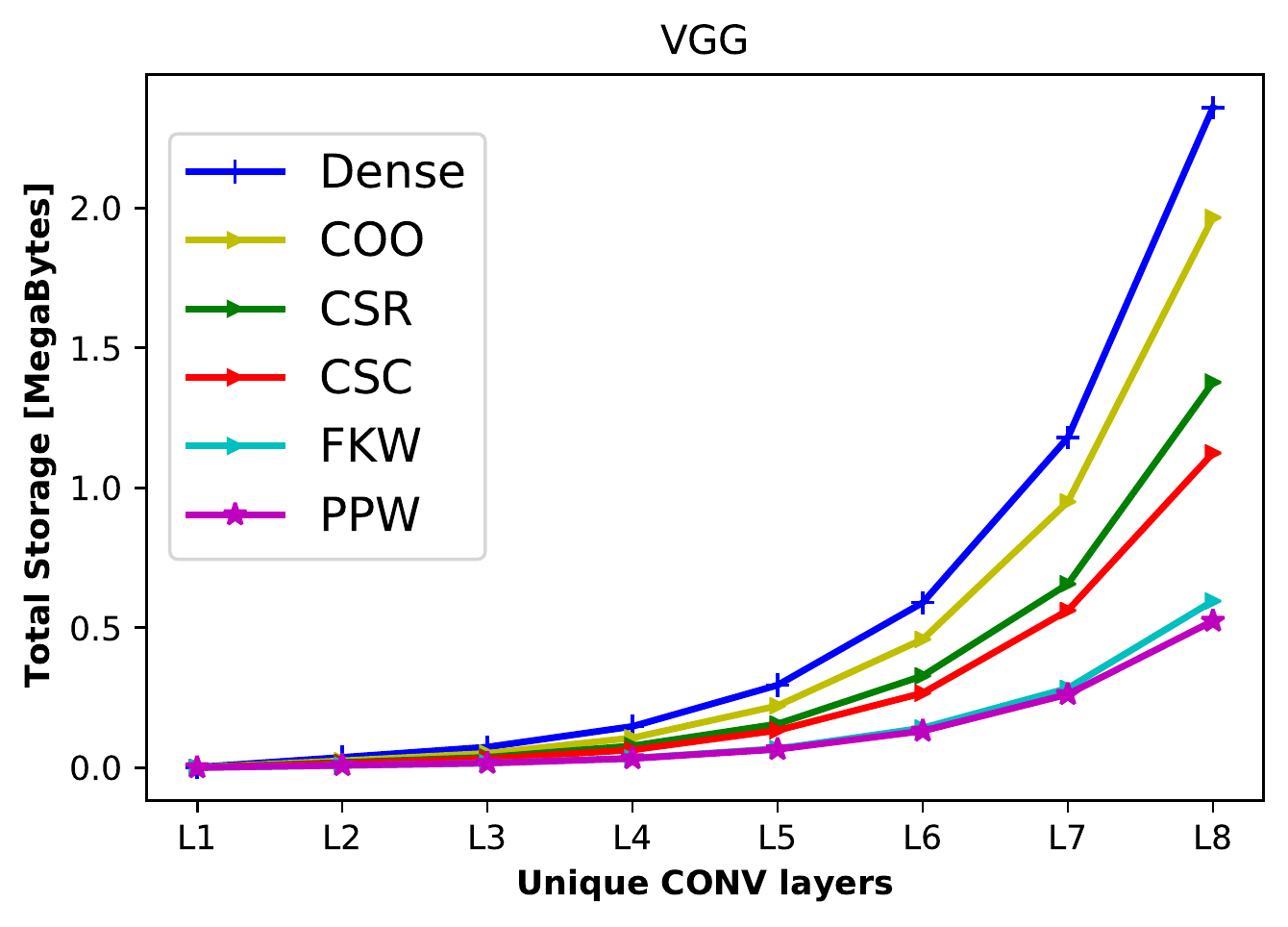} 
    \end{minipage}\hfill
    \begin{minipage}{0.48\columnwidth}
        \centering
        \includegraphics[width=\textwidth]{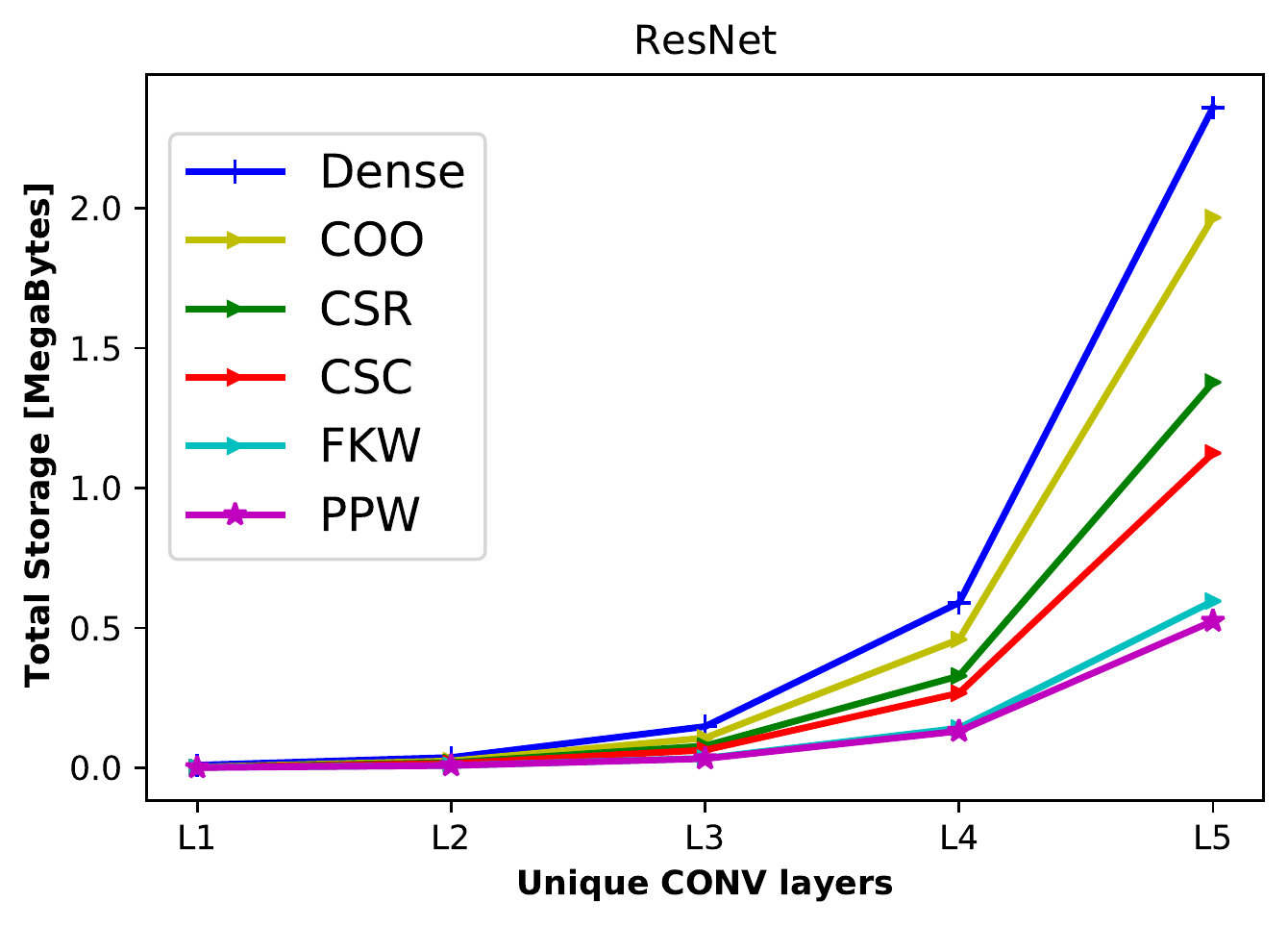}
    \end{minipage}
    \caption{Total (Weight + Index) storage comparison.}
    \label{fig:tot-storage}
\end{figure}

\begin{figure}[t]
\vspace{-10pt}
    \centering
    \begin{minipage}{0.48\columnwidth}
        \centering
        \includegraphics[width=\textwidth]{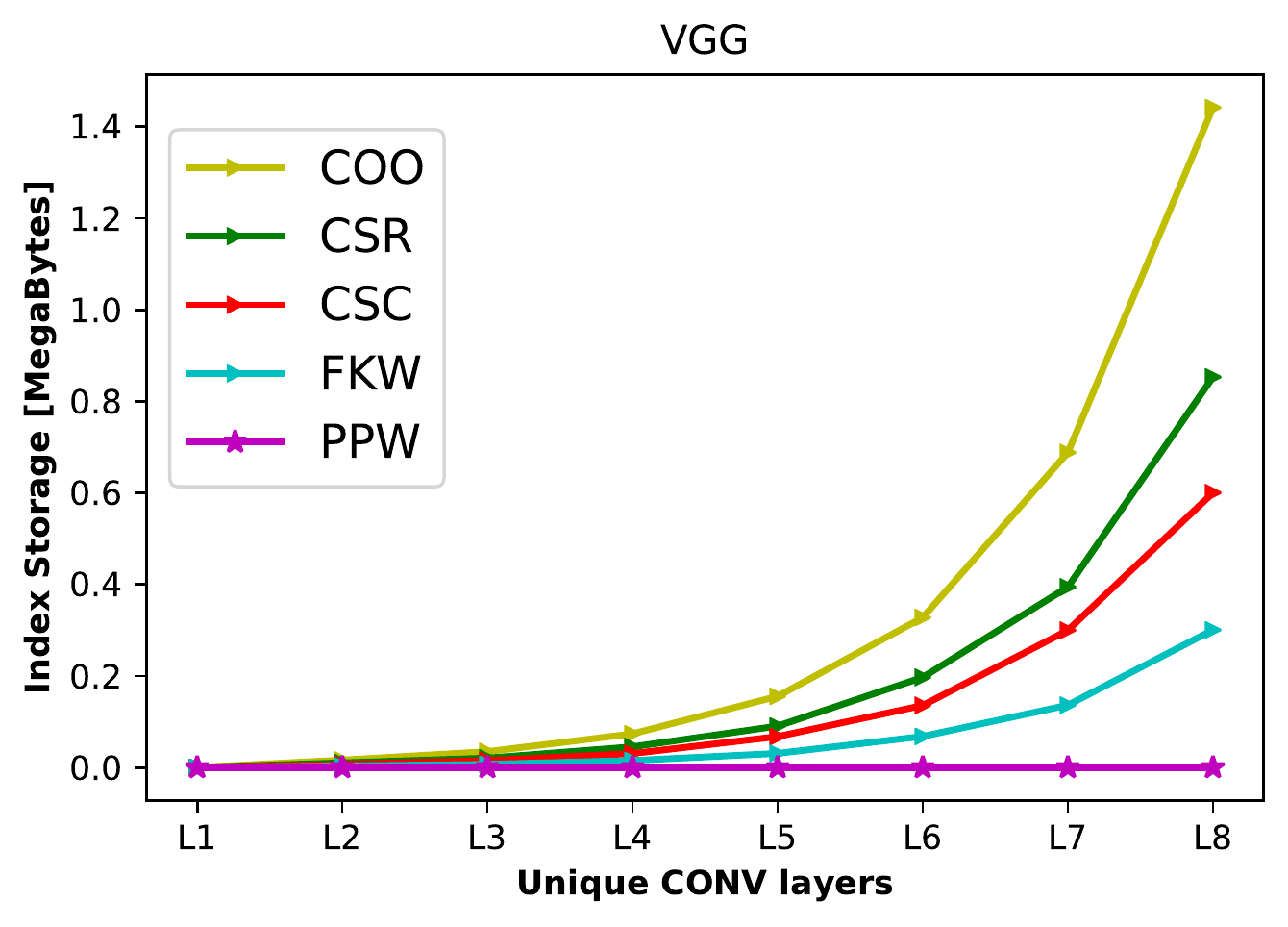} 
    \end{minipage}\hfill
    \begin{minipage}{0.48\columnwidth}
        \centering
        \includegraphics[width=\textwidth]{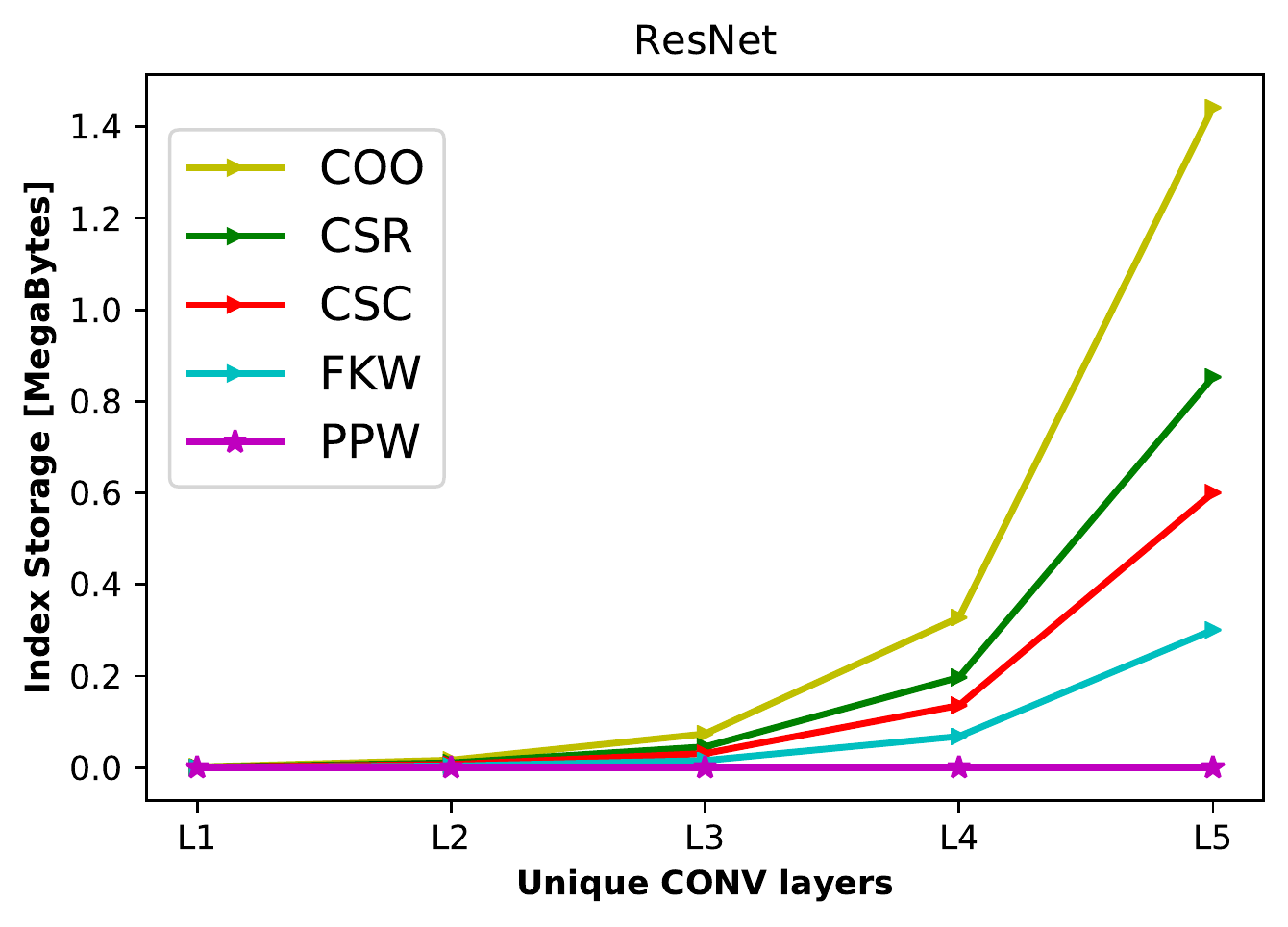}
    \end{minipage}
    \caption{Index storage comparison.}
    \label{fig:idx-storage}
\end{figure}

Fig. \ref{fig:tot-storage} shows that PPW compresses over 77.8\% (4.5x), 72.8\% (3.67x), 61.2\% (2.57x), 52.9\% (2.12x), 10.5\% (1.12x) of the total storage requirement compared to dense, COO, CSR, CSC, and FKW, respectively over unique convolutional layers in VGG16. Note that layers 8-10 and 11-13 in VGG16 has the same weight matrix size and are represented by L7 and L8 in our figures. Similar selection has been adopted for ResNet18. Dense model represents the non-pruned baseline model.

Fig. \ref{fig:idx-storage} illustrates that PPW format requires 22044x less indexing storage even compared to the FKW format, which is the most competitive. As seen on the graph, PPW enjoys constant storage requirement of small amount of bits %\textbf{35 bits} 
across different convolutional layers in VGG, while others grow on the order of \textbf{MegaBytes}. Similar effects are shown in selected convolutional layers of ResNet18, where PPW consistently outperforms total storage while remaining near zero-valued for indexing storage. % This effectively is the same as the dense format, where 0 bit is required for auxiliary buffers. 

\begin{figure}[b]
    \centering
    \begin{minipage}{0.48\columnwidth}
        \centering
        \includegraphics[width=\textwidth]{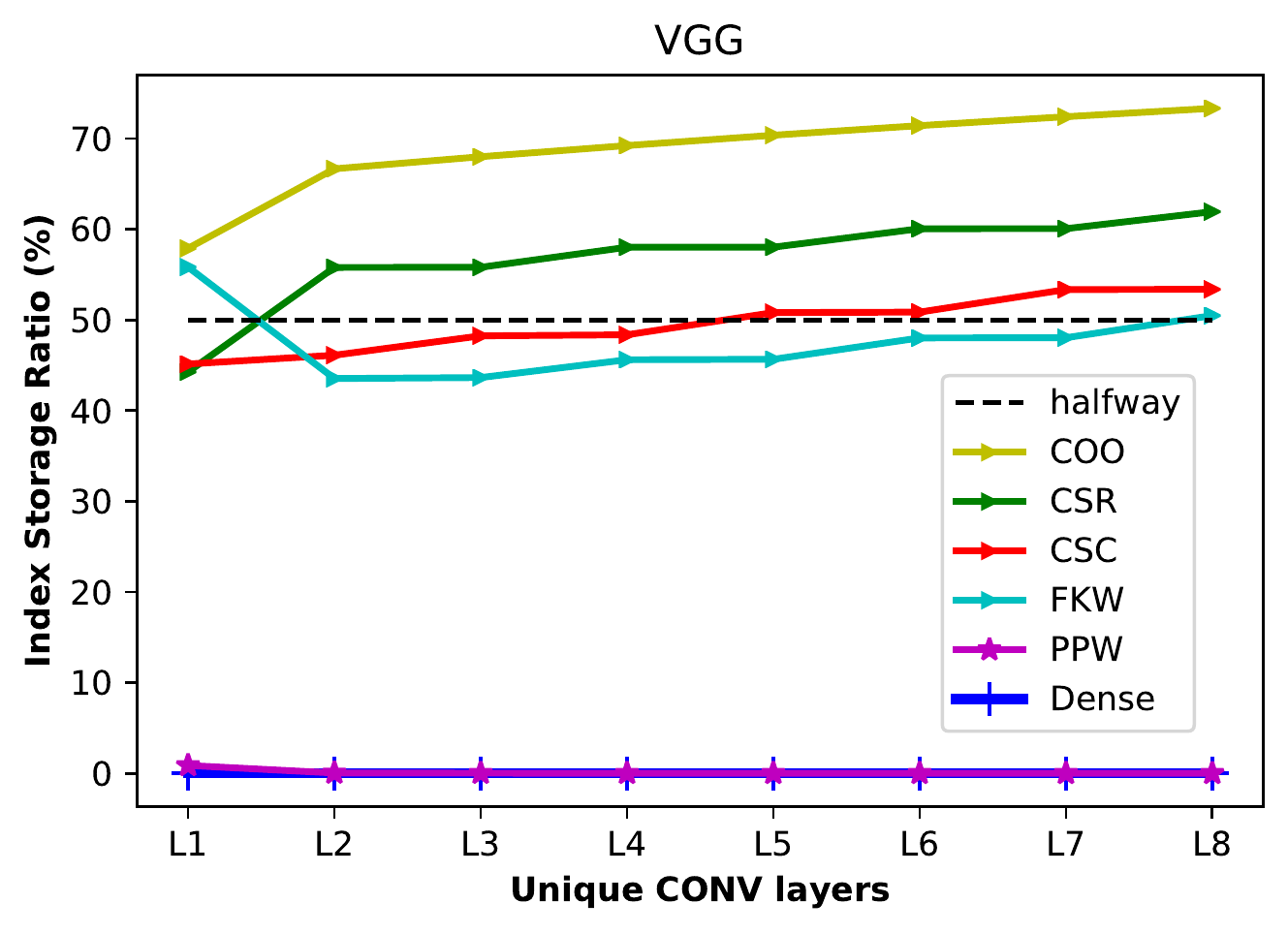} 
    \end{minipage}\hfill
    \begin{minipage}{0.48\columnwidth}
        \centering
        \includegraphics[width=\textwidth]{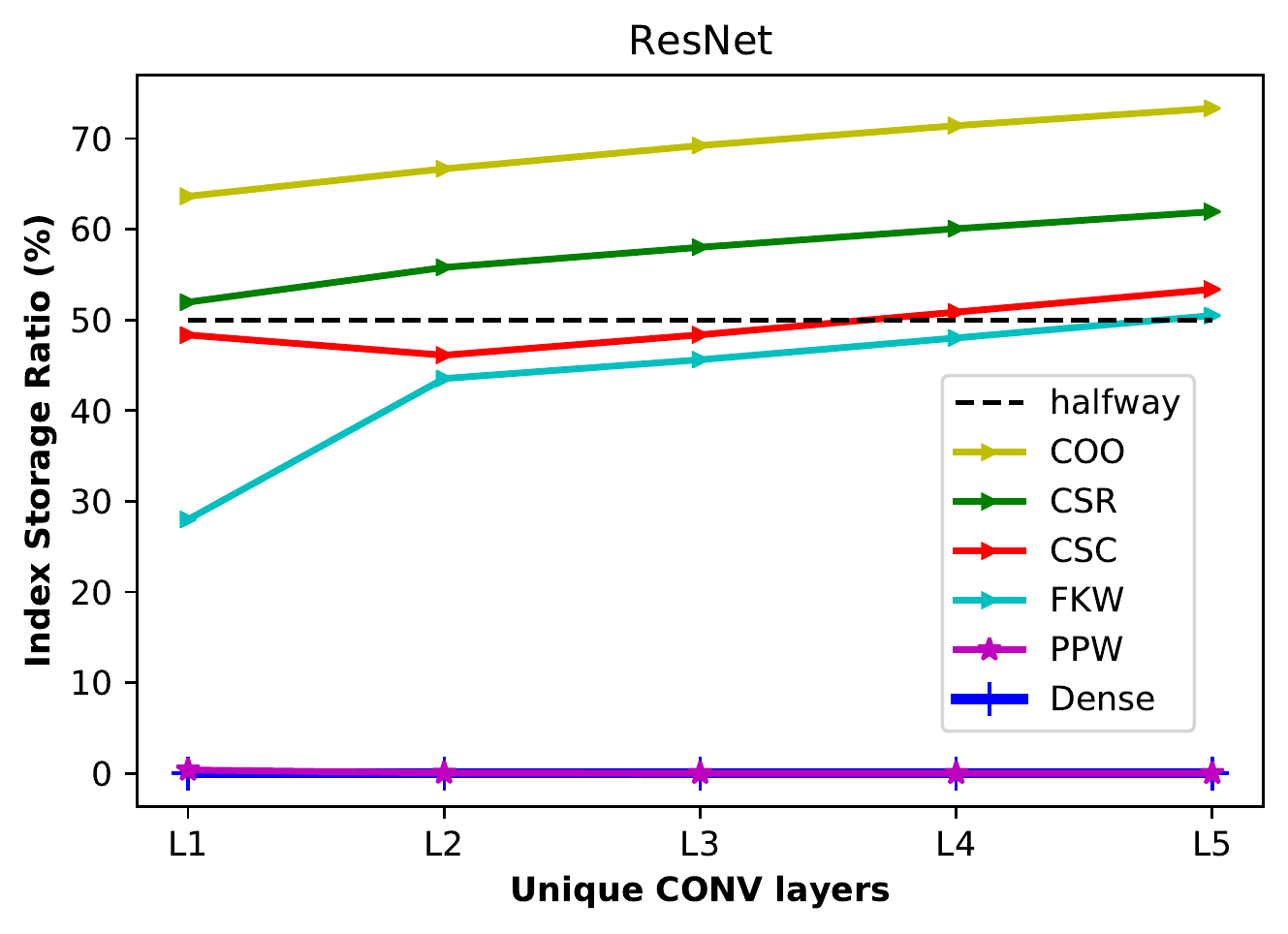}
    \end{minipage}
    \caption{Percent Storage for unique layers.}
    \label{fig:ind-storage-ratios}
\end{figure}

For larger convolutional layers (the later layers) with many irregular weights, more space is dedicated to store indexing buffers than the actual weights (see Fig. \ref{fig:ind-storage-ratios}). The halfway point (50\%) of the total storage is marked with a dotted line, and we can observe that most storage formats easily exceed this threshold. This also shows that the marginal increase in indexing storage is greater than that of weight storage. For example, COO steadily uses higher proportions of storage for indexing with larger convolutional layers, from the low point of 58\% in the smallest layer to 73\% in the largest layer.

\begin{figure}[t]
    \centering
    \begin{minipage}{0.48\columnwidth}
        \centering
        \includegraphics[width=\textwidth]{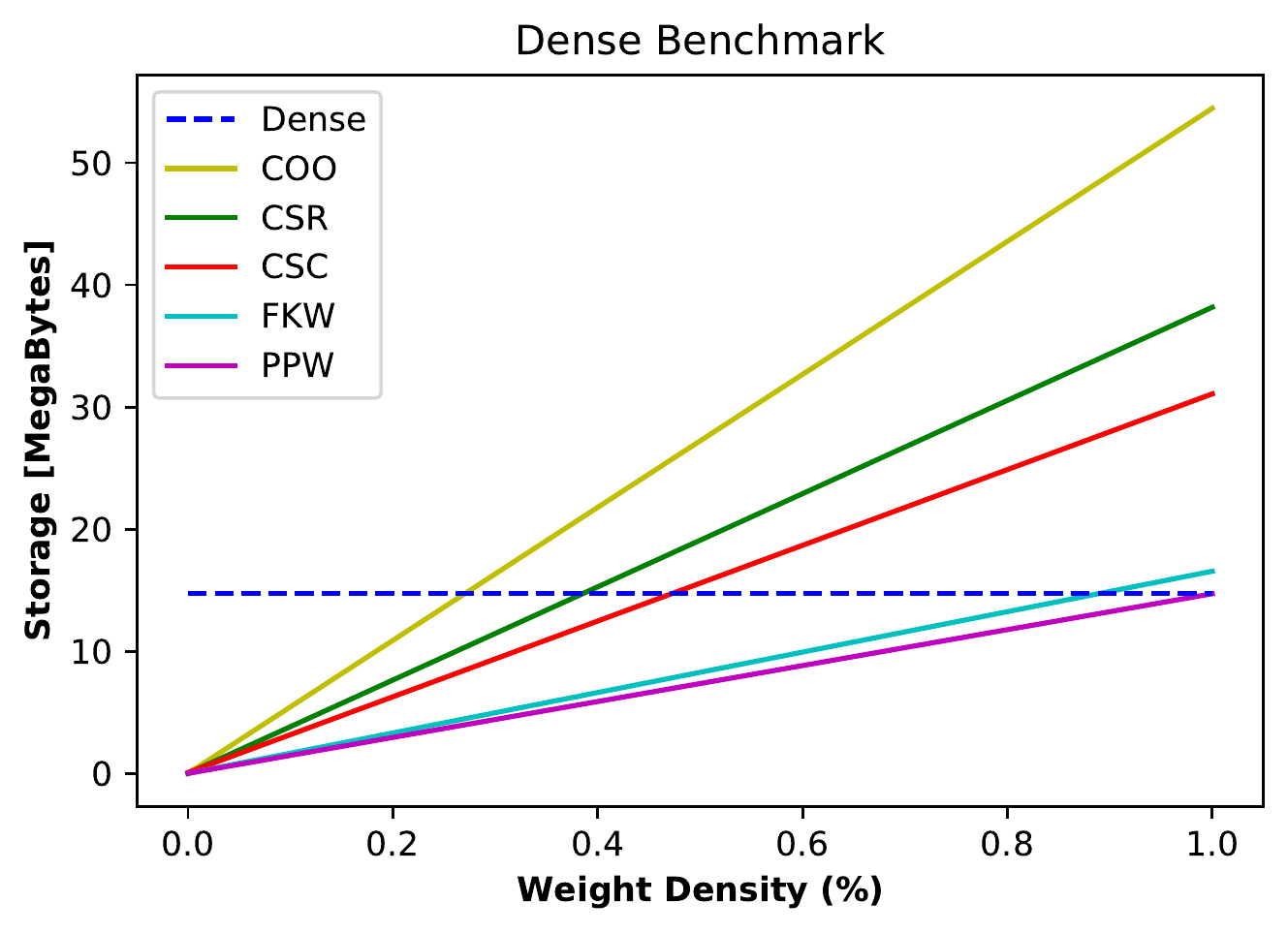}
    \end{minipage}\hfill
    \begin{minipage}{0.48\columnwidth}
        \centering
        \includegraphics[width=\textwidth]{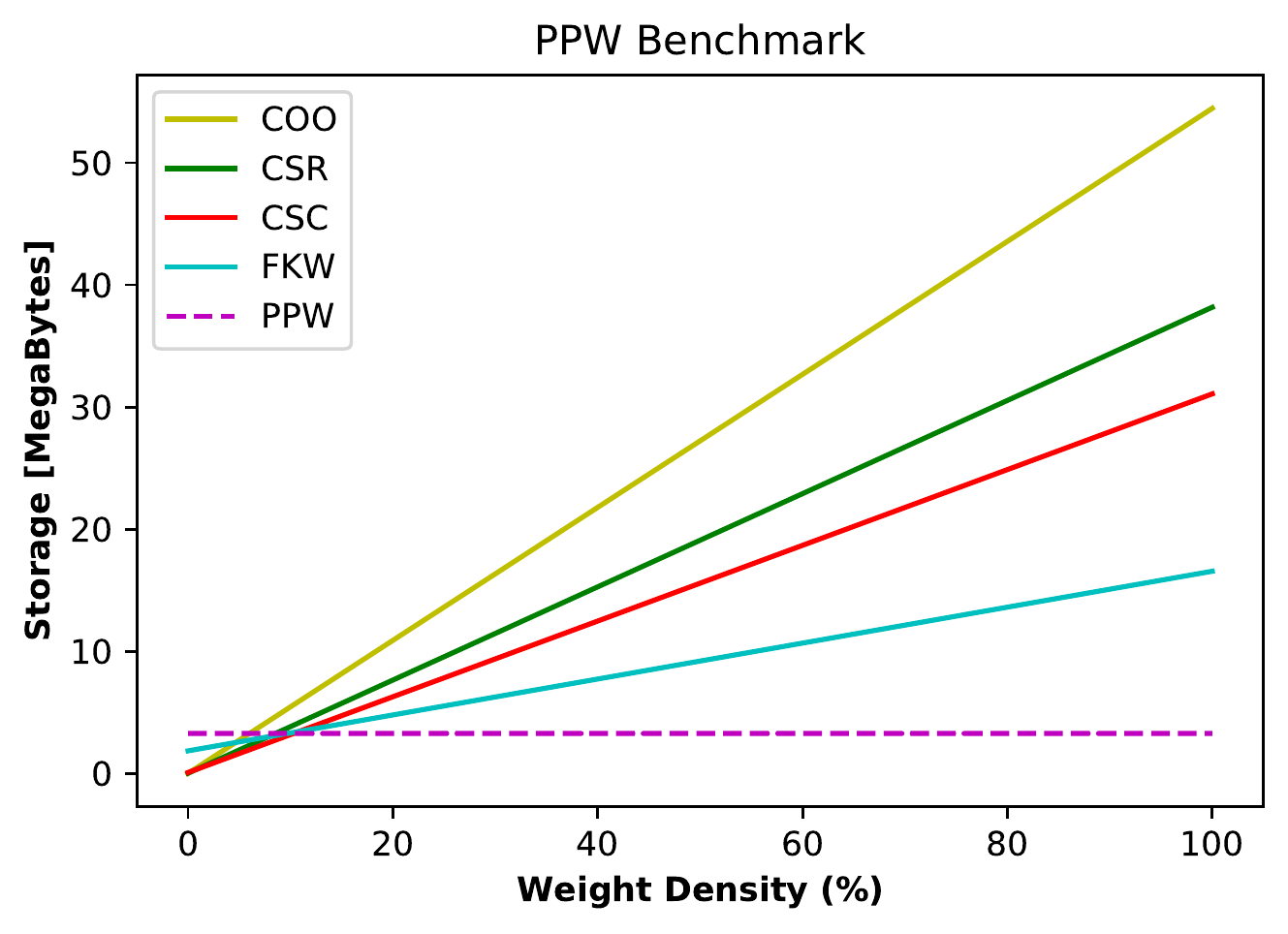} 
    \end{minipage}
    \caption{\small Benchmarking effective sparsity threshold for VGG16.}
    \label{fig:bench}
    %\addtolength{}{\belowcaptionskip}{1pt}
\end{figure}

To conduct a comparative analysis on a single storage format against others, we benchmark the \emph{effective sparsity threshold}, which is defined as the minimum sparsity rate that the weight format must achieve in order to realize a lower total storage requirement. This attempts to answer the question: given the pruning ratio and the total storage used for the benchmarked format, what is the minimum sparsity that the other formats must achieve in order to save more storage?

Our results from the left plot of Fig. \ref{fig:bench} show that most traditional sparse weight storage formats have a relatively harsh constraint on the sparsity requirement, with COO requiring only 22.2\% of weights to be unpruned. In other words, it requires 77.8\% of the weights to be pruned in order to begin saving more storage than the baseline dense format. 

%Pattern-based storage formats, thanks to more effective indexing units, found to be useful across all compression ratios.

% For a sparse storage format to be effective against the dense baseline (dotted line), the storage requirement needs to remain under the dotted line. This is because storage directly relates to hardware performance. We observe that the storage requirements exceed that of a dense format (without pruning) after some weight density threshold, i.e., \% of the weights remaining nonzero (see Fig. \ref{fig:bench_baseline}). The lower this threshold is, the higher compression ratio the format must achieve during the model pruning stage to attain the expected hardware performance from pruning. This imposes a constraint that cannot easily be met, as model compression is effective when the minimal accuracy degradation is achieved. 

We also benchmark PPW and observe as shown on the right of Fig. \ref{fig:bench}  that the highly compact format of PPW enforces a strict effective sparsity threshold on all other weight storage formats. The effective sparsity threshold for FKW, the most competitive format, is 90\%, which means that FKW requires the compression ratio to be at least 10x to begin saving more storage than PPW. Yet, reference \cite{patdnn} reports an 8x pruning rate, suggesting that the PPW format saves more storage under similar network accuracy. FKW could employ harsher pruning and achieve more than 10x pruning rate to realize lower storage, but this would nontrivially sacrifice the network accuracy under 91\% which is more than the acceptable 1\% degradation range. 

\subsection{Hardware Utilization and Energy Efficiency Comparisons}

\begin{table}[b]
  \caption{\small Hardware Utilization for VGG16 on CIFAR-10.}
  \label{table:resource-util}
  \resizebox{\columnwidth}{!}{%
  %\scriptsize%\footnotesize%\tiny%\scriptsize% % text size of table content
  \centering % center the table
  \begin{tabular}{|c|c|c|c|c|} % alignment of each column data
  \toprule[\heavyrulewidth]
   \textbf{Hardware resource}  & DSP48E & LUT & BRAM\_18K & Frequency (MHz) \\
  \midrule
  \textbf{Usage in our architecture} & 1038 (15\%) &  115290 (10\%) & 512 (12\%) &  342\\
   \midrule
  \textbf{Baseline architecture}$^*$ & 1038 (15\%) &  115290 (10\%) & 2942 (68\%) &  342\\
  \midrule
  \end{tabular}}
  \begin{flushleft}
    $^*${\scriptsize the baseline architecture is used for handling the dense format.} \\
\end{flushleft}
\end{table}

In this section, we evaluate the aforementioned sparsity storage formats in the FPGA platform.
First, the hardware utilization of the proposed accelerator tailored to SPS dataflow for running VGG16 on CIFAR-10 dataset is reported in Table \ref{table:resource-util}. The baseline architecture is similar to the architecture shown in Fig. \ref{fig:arch} while removing IMU and weight index buffer. As shown in the table, when employing the accelerator design discussed in Section \ref{sec:exp}, periodic pattern-based pruning that eliminates 77.8\% of the weights stored in the BRAM alongside with the PPW storage format that requires minimal indexing support in hardware leads to efficient usage of hardware resources in the FPGA. 

\begin{figure}[t]
    \centering
    \includegraphics[width=0.6\columnwidth]{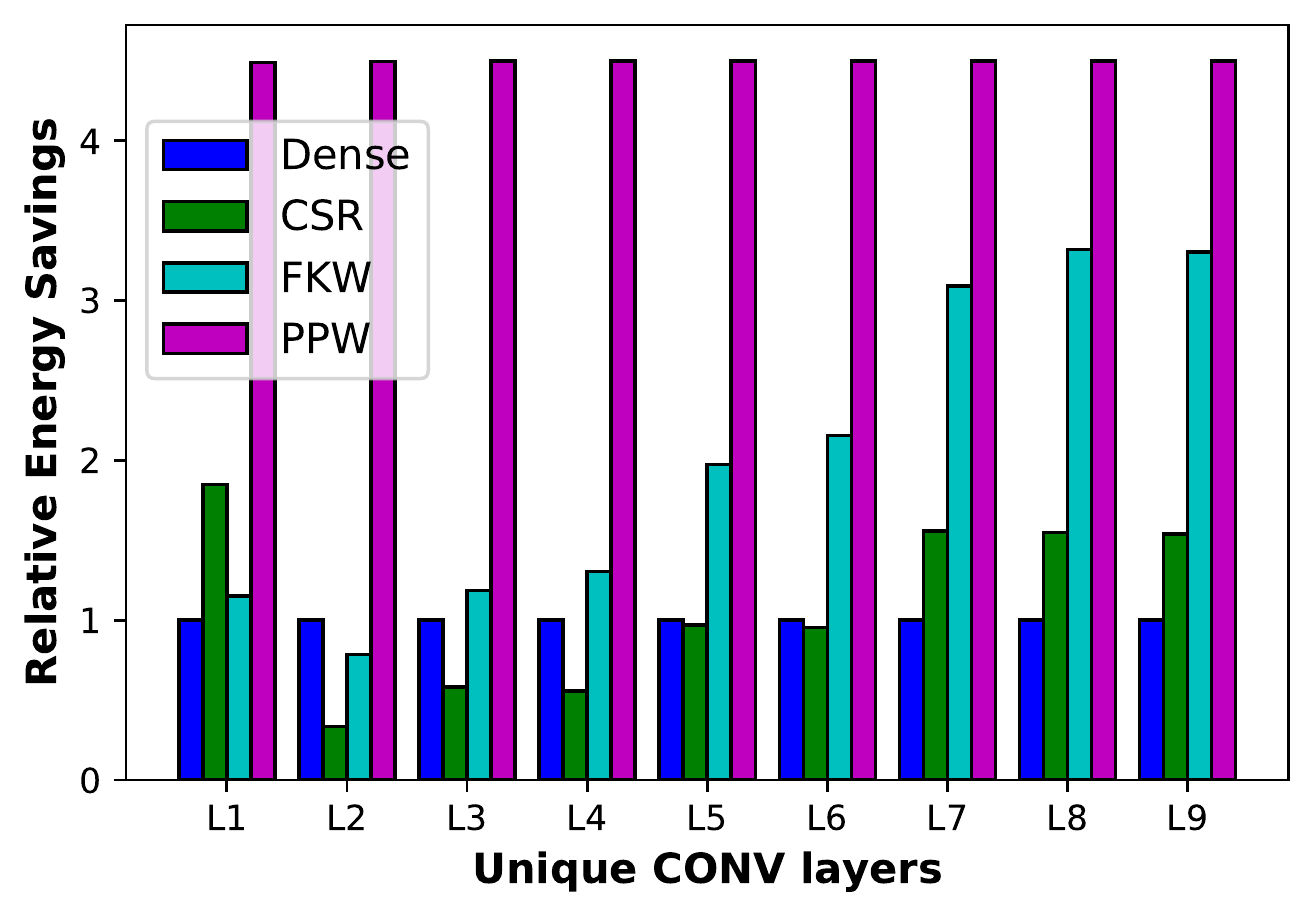}
    \caption{Energy Savings over dense baseline for VGG16.}
    \label{fig:energy-savings}
\end{figure}

Next, we evaluate the energy efficiency of our proposed architecture and dataflow compared to other formats. CSR and FKW are implemented as they are the competitive formats that exist today. The relative energy savings is reported in Fig. \ref{fig:energy-savings}, normalized with the dense baseline architecture. Our PPW format executed by the SPS dataflow achieves 4.49$\times$ energy savings over the dense baseline, while CSR acehives 1.4$\times$ and FKW achieves 3.1$\times$. To understand the energy savings, we classify the resources of energy cost in four ways: 1) bringing in weights 2) running MAC operations 3) read/write from/to weight index buffers and 4) data reordering cost for pattern-based dataflows such as the FKW. The first two costs are very similar as they're directly proportional to the the number of weights being moved around the hardware (thus modeled by the weight density). However, the third cost poses a nontrivial challenge to CSR, while FKW and PPW are relatively immune to the indexing overhead that occurs while supporting the MAC operation. This follows suit in Fig. \ref{fig:idx-storage}. 4) is unique to pattern-based formats, where FKW pays the cost of reordering the number of output channels that have been mixed during the compiler optimization. Such indexing overhead occurs in every layer, as the output feature map is the input feature map of the next layer, and the dataflow expects it to be in the natural, unmixed order. On the other hand, SPS dataflow's next layer reordering allows outputs to be grouped together without the need of data reordering, which eliminates the cost \#4. 

\section{Conclusion}\label{conc}
The SPS dataflow offers a novel hardware design approach afforded by periodic pattern-based sparsity, resulting in neural network weights with higher degrees of regularity and thus parallelism.
By avoiding excessive indexing costs with the compiler-hardware co-design approach, the SPS dataflow outperforms state-of-the-art sparisty formats in CNN accelerator designs targeting FPGAs.% devices.

\textbf{Acknowledgment:} This research is supported by a grant from the Software and Hardware Foundations program of the NSF.

\balance

\bibliographystyle{ACM-Reference-Format}
\bibliography{main}


\end{document}